\pdfoutput=1

\documentclass[11pt]{article}

\usepackage[]{ACL2023}

\usepackage{times}
\usepackage{latexsym}

\usepackage[T1]{fontenc}

\usepackage[utf8]{inputenc}

\usepackage{microtype}
\usepackage{multirow}
\usepackage{booktabs}
\usepackage{subcaption}
\usepackage{graphicx}
\usepackage{xurl}  
\usepackage{enumitem}  
\usepackage{makecell}  
\usepackage{amssymb, amsmath}
\newcolumntype{H}{>{\setbox0=\hbox\bgroup}c<{\egroup}@{}}

\title{
Causes and Cures for Interference in Multilingual Translation
}

\author{
Uri Shaham$^\tau$ \qquad Maha Elbayad$^\mu$ \qquad Vedanuj Goswami$^\mu$ \\
\textbf{Omer Levy}$^{\tau\mu}$ \qquad \textbf{Shruti Bhosale}$^\mu$ \\
\\
$^\tau$ The Blavatnik School of Computer Science, Tel Aviv University \\
$^\mu$ Meta AI 
}

\begin{document}
\maketitle

\begin{abstract}
Multilingual machine translation models can benefit from synergy between different language pairs, but also suffer from interference.
While there is a growing number of sophisticated methods that aim to eliminate interference, our understanding of interference as a phenomenon is still limited.
This work identifies the main factors that contribute to interference in multilingual machine translation.
Through systematic experimentation, we find that interference (or synergy) are primarily determined by model size, data size, and the proportion of each language pair within the total dataset.
We observe that substantial interference occurs mainly when the model is very small with respect to the available training data, and that using standard transformer configurations with less than one billion parameters largely alleviates interference and promotes synergy.
Moreover, we show that tuning the sampling temperature to control the proportion of each language pair in the data is key to balancing the amount of interference between low and high resource language pairs effectively, and can lead to superior performance overall.
\end{abstract}

\section{Introduction}

Multilingual machine translation models can benefit from transfer between different language pairs (\textit{synergy}), but may also suffer from \textit{interference} \cite{ha-etal-2016-toward,firat-etal-2016-multi,aharoni-etal-2019-massively,Arivazhagan2019MassivelyMN}. 
While there are methods to reduce interference and achieve better performance \cite{wang-etal-2020-balancing,kreutzer-etal-2021-bandits-dont,wang2021gradient}, such approaches are often compute intensive, and do not always work \cite{xin2022do}. 
In this work, we demonstrate that interference in multilingual translation largely occurs when the model is very small compared to the abundance of training data, 
and that the simple principled approach of enlarging the model and tuning the data sampling temperature provides a consistent solution to the interference problem that can even promote synergy.

This work methodically deduces the most simple ways of reducing interference in multilingual translation.
We begin by inquiring what are the dominant factors that may interfere with learning to translate a particular language pair of focus $s \rightarrow t$, in the context of learning a multilingual translation model with many different language pairs.
Controlled experiments show that besides model size and number of $s \rightarrow t$ training examples, the main factor that correlates with the level of interference is the proportion of \textit{focus pair} examples ($s \rightarrow t$) observed out of the \textit{total} number of examples (all language pairs) seen at each training step on average.
Surprisingly, aspects like language similarity or number of translation directions have a much smaller effect.

In model and data scaling experiments, we observe that interference mainly occurs in extreme parameter poverty, when the language pair of focus is data-rich, but has to ``share'' a crowded parameter space with large quantities of other data.
Enlarging the model to standard model sizes in machine translation literature alleviates interference and even facilitates synergy.
For context, given a language pair of 15M sentence pairs that accounts for 20\% of the total training data (75M), we observe severe levels of interference with 11M- and 44M-parameter transformers, but no interference when scaling the model to 176M parameters (the ``big'' model of \citet{vaswani2017attention}) and significant synergy with 705M parameters.
Interestingly, when the model is large enough, we find that increasing the amount of non-focus data to a certain point can further increase synergy.

Finally, given the evidence that data sizes and ratios strongly correlate with interference,
we experiment with a natural lever that controls the proportion of each dataset in the overall mix in the simplest way: sampling temperature.
Indeed, we find that calibrating the distribution of language pairs via temperature can substantially reduce the amount of interference in both high- and low-resource language pairs.
Our results demonstrate the importance of tuning the temperature hyperparameter in multitask training, and suggest that previously reported accounts of severe interference in multilingual translation models might stem from suboptimal hyperparameter configurations.

\section{Measuring Interference}
\label{sec:interference}

We assume a common multilingual translation setup that involves $L$ language pairs $s \rightarrow t$,
where the source is always the same language $s$ (English), and the target language $t$ varies (English-to-many), or vice versa (many-to-English). The overall training data is a union of these training subsets, we note their sizes by $D_{s \rightarrow t}$. Sampling a training example $x$ follows the distribution: 
\begin{equation}
P(x \in s \rightarrow t) \propto \bigg(\frac{D_{s \rightarrow t}}{ \sum\limits_{s^{\prime}, t^{\prime}} D_{s^{\prime} \rightarrow t^{\prime}}}\bigg)^{\frac{1}{T}}
\label{eq:proportion}
\end{equation}
Where $T$ is the temperature hyperparameter \cite{devlin-etal-2019-bert,Arivazhagan2019MassivelyMN}. $T=1$ maintains the original data proportions,  $ 0 < T < 1$ starves low resource language pairs, and $ T > 1$ increases their representation in the training distribution.
We mostly focus on the English-to-many setting in which interference is more apparent.\footnote{Section~\ref{sec:scale} also includes many-to-English experiments, where we observe higher levels of synergy.}

We define interference as a negative interaction between different translation directions in a multilingual translation model. 
It is measured for a specific translation direction $s \rightarrow t$  by the relative difference in performance (test-set cross-entropy loss) between a bilingual model trained to translate only from $s$ to $t$ ($\mathcal{L}^{\text{bi}}_{s \rightarrow t}$) and a multilingual counterpart that is trained to translate other additional directions ($\mathcal{L}^{\text{multi}}_{s \rightarrow t}$): 
\begin{equation}
\mathcal{I}_{s \rightarrow t} = \frac{\mathcal{L}^{\text{bi}}_{s \rightarrow t} - \mathcal{L}^{\text{multi}}_{s \rightarrow t}}{ \mathcal{L}^{\text{bi}}_{s \rightarrow t}}
\end{equation}
Negative values of $\mathcal{I}_{s \rightarrow t}$ indicate interference, while positive values indicate synergy.

\section{Experimental Setup}

\paragraph{Models}
We train encoder-decoder Transformer \cite{vaswani2017attention} models of 4 different sizes throughout our experiments. We use the original\footnote{With pre-layer normalization and a shared embedding matrix across the encoder input, decoder input, and decoder output \cite{press-wolf-2017-using}.}
transformer-base and transformer-big variants, as well as a smaller and a larger versions by adjusting the width of the architecture (Table~\ref{tab:arch}).

\paragraph{Data}
We use the multilingual benchmark introduced by \citet{siddhant-etal-2020-leveraging} based on WMT data.
This benchmark includes a diverse set of 15 languages, each paired with English. The number of training examples is also diverse, ranging from 155K sentence pairs in Gujarati to 51M examples in Czech.\footnote{Note that \citet{siddhant-etal-2020-leveraging} only uses 11K pairs in Gujarati whereas we use the additional training data recommended by the WMT'19 shared task (\url{https://statmt.org/wmt19/translation-task.html}).}
Table \ref{tab:data} provides additional dataset statistics.

\paragraph{Tokenization}
We build a shared vocabulary of 64K BPE tokens with sentencepiece \cite{kudo-richardson-2018-sentencepiece} using a sampling temperature of 5 to increase the lower resource languages' representation. We use this vocabulary for all our experiments.
We also add language ID tokens to our vocabulary, which are prepended to each source and target sequence to indicate the target language \citep{johnson-etal-2017-googles}.

\paragraph{Training}
We use Fairseq \cite{ott-etal-2019-fairseq} to train transformer models with the Adam optimizer \cite{DBLP:journals/corr/KingmaB14} for up to 100K steps, with a dropout rate of 0.1, inverse square root learning rate schedule up to a maximum of 0.004, 8K warmup steps, and a batch size of 256K tokens.
We choose the best checkpoint according to the average validation loss of all language pairs.

\begin{table}[t]
\small
\centering
\begin{tabular}{@{}lrrrr@{}}
\toprule
\textbf{Size} & \textbf{Hidden} & \textbf{FFN} &  \textbf{Attn Heads} & \textbf{Params}  \\
\midrule
XS & 256 & 1024 & 4 & 11M   \\
S & 512 & 2048 & 8 & 44M   \\
M & 1024 & 4096 & 16 & 176M   \\
L & 2048 & 8192 & 32 & 704M   \\
\bottomrule
\end{tabular}
\caption{Model sizes used in our experiments. Each model has 6 encoder and 6 decoder layers. We exclude the embeddings from the parameters count.}
\label{tab:arch}
\end{table}

\begin{table}[t]
\small
\centering
\begin{tabular}{@{}llrr@{}}
\toprule
\textbf{Language} & \textbf{ID} &  \textbf{\#Sentences (M)}  &  \textbf{Test Set}  \\
\midrule
Czech & cs & 51.769  & WMT18 \\
French & fr & 40.853 & WMT14 \\
Russian & ru & 38.492  & WMT19 \\
Chinese & zh & 25.987   & WMT19 \\
Spanish & es & 15.177   & WMT13 \\
Finnish & fi & 6.587  & WMT19 \\
German & de & 4.509  & WMT14 \\
Estonian & et & 2.176  & WMT18 \\
Latvian & lv & 0.638 & WMT17 \\
Lithuanian & lt & 0.631  & WMT19 \\
Romanian & ro & 0.610 & WMT16 \\
Hindi & hi & 0.306 & WMT14 \\
Kazakh & kk & 0.224  & WMT19 \\
Turkish & tr & 0.207 & WMT18 \\
Gujarati & gu & 0.156 & WMT19 \\
\bottomrule
\end{tabular}
\caption{Languages from the WMT-based benchmark of \citet{siddhant-etal-2020-leveraging}, along with the number of sentence pairs in the training set, and the source of the test set. All languages are paired with English (en).
}
\label{tab:data}
\end{table}

\section{What Impacts Interference in Multilingual Translation?}

We consider 5 factors that may potentially impact the performance of a given language pair $s \rightarrow t$ in the multilingual translation setting:
\setlist{nolistsep}
\begin{enumerate}[noitemsep,label={(\arabic*)}]
    \item Model size
    \item Training data size of $s \rightarrow t $,  $D_{s \rightarrow t}$
    \item Proportion of $s \rightarrow t$ examples observed during training $P(x 
\in s \rightarrow t)$
    \item Total number of languages $L$
    \item Similarity between $s \rightarrow t$ and other pairs\footnote{While the other factors can be exactly quantified, it is not immediately clear how to measure language similarity. In our experiments, we use a phylogenetic interpretation of language similarity within the set of languages available in our dataset.} 
\end{enumerate}
In the experiments we describe next, we provide empirical evidence that indicate the last two factors do not actually have a significant effect on the level of interference, and can therefore be pruned away.
Subsequent experiments reveal that interference is indeed a function of model size, data size, and data proportion.
Most striking is the fact that, across various data settings, enlarging the model to standard sizes consistently alleviates interference and may even promote synergy.

\subsection{Does Language Similarity Matter?}
\label{sec:lang_sim}

\begin{figure*}[t]
\centering
\begin{subfigure}[b]{1.0\textwidth}
\centering
   \includegraphics[width=1\linewidth]{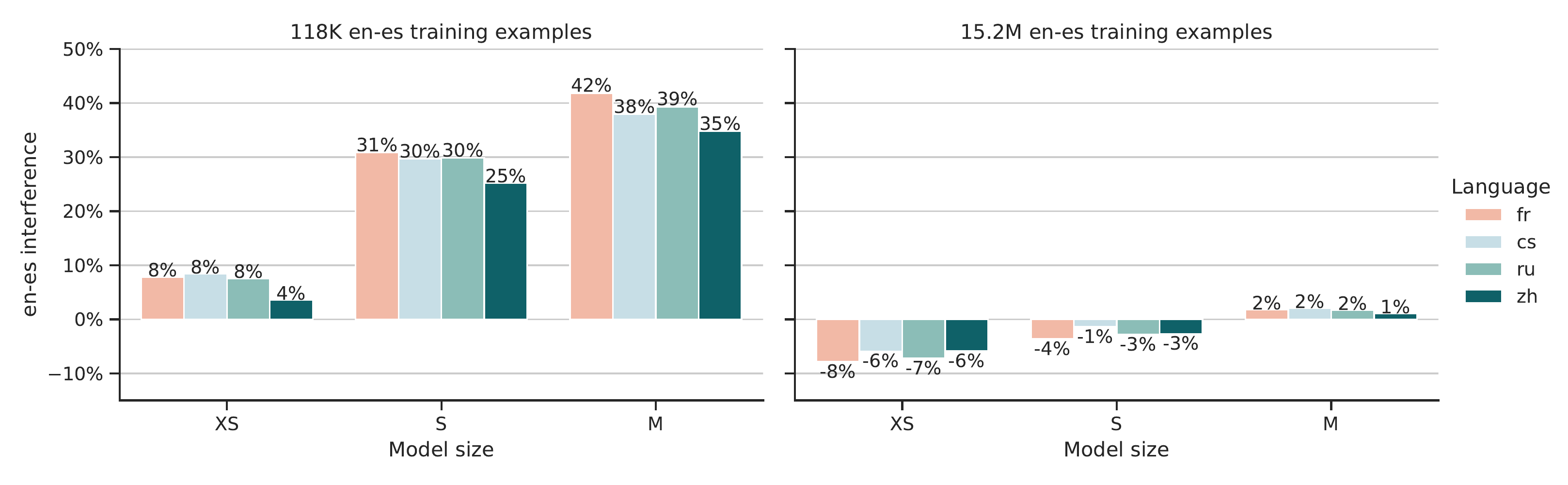}
   \caption{Models trained with 118K (left) and 15.2M (right) en-es examples together with 15.2M examples of en-xx. }
   \label{fig:lang_sim_es} 
\end{subfigure}
\begin{subfigure}[b]{1.0\textwidth}
\centering
   \includegraphics[width=1\linewidth]{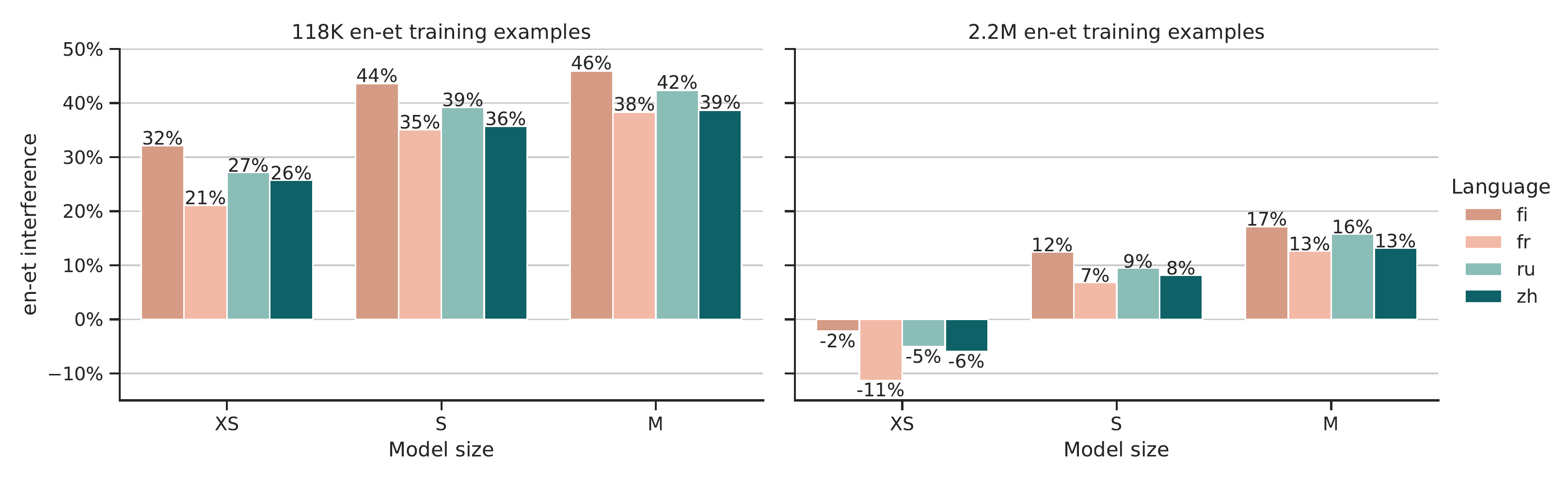}
   \caption{Models trained with 118K (left) and 2.2M (right) en-et examples together with 6.6M examples of en-xx.}
   \label{fig:lang_sim_et}
\end{subfigure}
 \label{fig:lang_sim}
\caption{Interference of models trained with en-es (a) or en-et (b) as low resource languages (left) and using their full training sets (right) together with one other language. Positive values indicate synergy, i.e. the focus language (es/et) loss of a trilingual model is lower (better) compared to its bilingual model baseline. Similarly, negative values indicate interference.}
\end{figure*}

Intuitively, data from languages that humans perceive as similar (e.g. languages that have some degree of mutual intelligibility, exhibit similar linguistic properties, or have shared vocabularies) should have a more positive effect on translation quality comparing to data from distinct languages \cite{lin-etal-2019-choosing,wang-etal-2020-negative}.
To test this, we fix a \textit{focus} language, and train \textit{trilingual} models to translate from English to two languages, the focus language and an additional \textit{interfering} language.
We then look at interference trends as we vary the interfering language while controlling the amount of training data for each language pair.

\paragraph{Setup}
We run two sets of experiments, one with Spanish (es, 15.2M parallel sentences) as the focus language, and another with Estonian (et, 2.2M examples).
For each focus language, we select one of four interfering languages; Spanish is paired with French,\footnote{Spanish and French are Western Romance languages.} Czech, Russian, and Chinese, while Estonian is paired with Finnish,\footnote{Estonian and Finnish are Balto-Finnic languages.} French, Russian, and Chinese.
To control the effects of data size in the English-Spanish experiments, we randomly sample 15.2M examples from each interfering language pair, making the ratio between focus and interfering languages 1:1.
Similarly, in the English-Estonian experiments, we sample 6.6M examples from each interfering language to create a data ratio of 1:3.
We also conduct similar experiments when we use only 118K focus language examples, to see the trends when the focus language pair is extremely low resource.\footnote{118K sentence pairs is 128th of the English-Spanish training set. It is approximately equivalent to translating 30 novels.}
Table~\ref{tab:lang_sim} provides an overview of the language similarity experiments.

\begin{table}[!t]
\small
\centering
\begin{tabular}{@{}lrlr@{}}
\toprule
\multicolumn{2}{@{}l}{\textbf{Focus}} & \multicolumn{2}{l@{}}{\textbf{Other}} \\
\textbf{Language} & \textbf{\#Examples} &  \textbf{Language}  & \textbf{\#Examples} \\
\midrule
es & 15.177M & fr$\star$/cs/ru/zh & 15.177M \\
es & 0.118M & fr$\star$/cs/ru/zh & 15.177M \\
\midrule
et & 2.176M & fi$\star$/fr/ru/zh & 6.587M \\
et & 0.118M & fi$\star$/fr/ru/zh & 6.587M \\
\bottomrule
\end{tabular}
\caption{Trilingual models for experiments on the impact of language similarity on interference. The most similar language to the focus language is noted with $\star$.
}\label{tab:lang_sim}
\end{table}

\paragraph{Results}
Figure \ref{fig:lang_sim_es} shows the interference rate for every model size when Spanish has only 118K parallel examples (left) and when using the full English-Spanish dataset (right).
The variance in results somewhat correlates with language similarity when the dataset is very small, which aligns with previous work \cite{lin-etal-2019-choosing}; French seems to help Spanish more than other languages when the model is big enough, while Chinese helps less.
However, when training with the full dataset, the differences between other languages diminish for all model sizes.  Concurrently, \citet{fernandes2023scaling} also found no significant difference for using French or Chinese as a third language combined with English-German in a very high resource setting (600M examples per language pair).

We observe similar trends when Estonian is the focus language. Figure~\ref{fig:lang_sim_et} shows that when Estonian only has 118K training examples, combining with Finnish data seems to have some positive effect. However, this effect also shrinks when using all of the English-Estonian train set (only 2.2M examples, compared to the 15.2M of English-Spanish) and a model that is not too small.\footnote{See Figure~\ref{fig:abs_bleu_lang_sim} in Appendix~\ref{sec:appendix_bleu} for the results of these experiments with absolute BLEU scores.}

\begin{figure*}[t]
\centering
\begin{subfigure}[b]{1.0\textwidth}
\centering
   \includegraphics[width=1\linewidth]{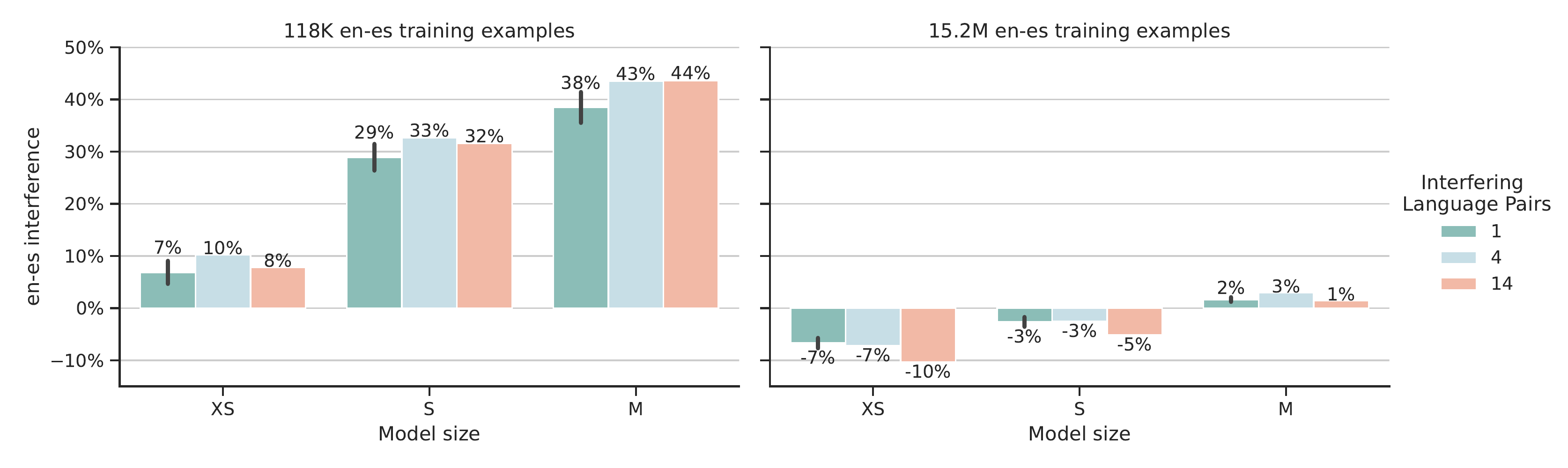}
   \caption{Models trained with 118K (left) and 15.2M (right) en-es training examples and 15.2M training examples for non-es languages. }
   \label{fig:num_tasks_es} 
\end{subfigure}

\begin{subfigure}[b]{1.0\textwidth}
\centering
   \includegraphics[width=1\linewidth]{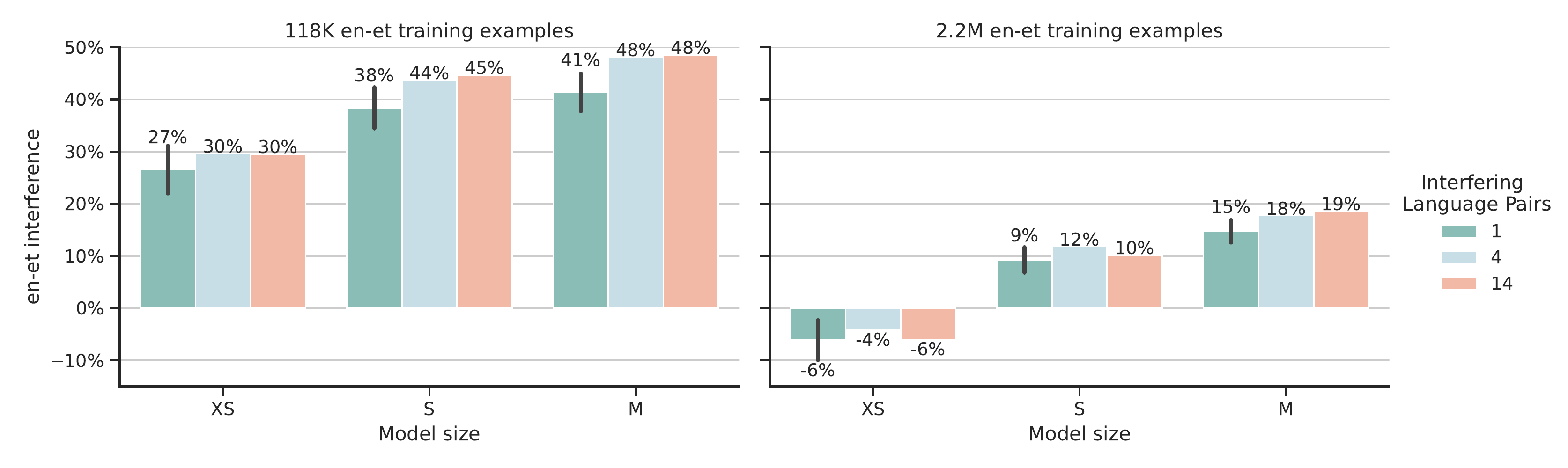}
   \caption{Models trained with 118K (left) and 2.2M (right) en-et training examples and 6.6M training examples for non-et languages.}
   \label{fig:num_tasks_et}
\end{subfigure}
\caption{en-es (a) and en-et (b) test interference of models trained with es (a) or et (b) as low resource languages (left) and using their full train sets (right) together with increasing number of languages, sharing a fixed budget of training examples. Positive values indicate synergy, i.e the focus language (es/et) loss of a multilingual model is lower (better) comparing to its bilingual model baseline. Similarly, negative values indicate interference. 
}
\label{fig:num_tasks}

\end{figure*}

\subsection{Does the Number of Languages Matter?}
\label{sec:num_tasks}

Do we get more interference when training with one interfering language pair or fourteen?
We train models with varying numbers of language pairs while controlling for the overall number of interfering examples.
We find that splitting the interfering data across more language pairs has a mild positive effect, which diminishes as the amount of focus-language data and/or model parameters scales up.

\begin{table}[!t]
\small
\centering
\begin{tabular}{@{}lrlr@{}}
\toprule
\multicolumn{2}{@{}l}{\textbf{Focus}} & \multicolumn{2}{l@{}}{\textbf{Other}} \\
\textbf{Language} & \textbf{\#Examples} &  \textbf{Languages}  & \textbf{\#Examples} \\
\midrule
\multirow{3}{*}{es}  & \multirow{3}{*}{15.177M} & cs/fr/ru/zh & 15.177M \\ 
  &  & cs+fr+ru+zh & 15.177M \\
  &  & cs+...+gu (14) & 15.177M \\
\midrule
\multirow{3}{*}{es}  & \multirow{3}{*}{0.118M} & cs/fr/ru/zh & 15.177M \\
  &  & cs+fr+ru+zh & 15.177M \\
  &  & cs+...+gu (14) & 15.177M \\
\midrule
\multirow{3}{*}{et}  & \multirow{3}{*}{2.176M} & fi/fr/ru/zh & 6.587M \\
  &  & fi+fr+ru+zh & 6.587M \\
  &  & cs+...+gu (14) & 6.587M \\
\midrule
\multirow{3}{*}{et}  & \multirow{3}{*}{0.118M} & fi/fr/ru/zh & 6.587M \\
  &  & fi+fr+ru+zh & 6.587M \\
  &  & cs+...+gu (14) & 6.587M \\
\bottomrule
\end{tabular}
\caption{Multilingual models for experiments on the impact of the number of other languages on interference. The trilingual model results are the average per focus language from Table~\ref{tab:lang_sim}.}
\label{tab:num_tasks}
\end{table}

\paragraph{Setup}
We train multilingual models on English-Spanish data alongside English to 1, 4, or 14 interfering languages.
The interfering data always sums up to a fixed 15.2M examples budget, distributed as evenly as possible among the different languages.\footnote{Some languages have less than 15.2M$/$14 (1.08M) examples. We use all of their training data, and divide the remaining budget evenly.}
We repeat these experiments when Estonian is the focus language and the interfering example budget is 6.6M. Table~\ref{tab:num_tasks} provides an overview of these experiments.

\paragraph{Results} 
Figure \ref{fig:num_tasks_es} shows that more than one interfering language pair somewhat helps when English-Spanish has few training examples, but this effect largely disappears in the full training set and with larger models.
We see similar trends for Estonian in Figure \ref{fig:num_tasks_et}, even though its full training set has only 2.2M examples.
This phenomenon might be related to the fact that when the data distribution is sharp (i.e. one high resource paired with one very low resource) there is not enough incentive for the model to pay attention to the focus language's identifier token, compared to when the distribution is much more uniform. 
This result also corroborates similar findings for pretrained multilingual models \cite{conneau-etal-2020-unsupervised}, although those experiments did not control the total quantity of data as in ours.\footnote{See Figure~\ref{fig:abs_bleu_num_tasks} in Appendix~\ref{sec:appendix_bleu} for the results of these experiments with absolute BLEU scores.}

\begin{figure*}[t]
\centering
\begin{subfigure}{.5\textwidth}
  \centering
  \includegraphics[width=\columnwidth]{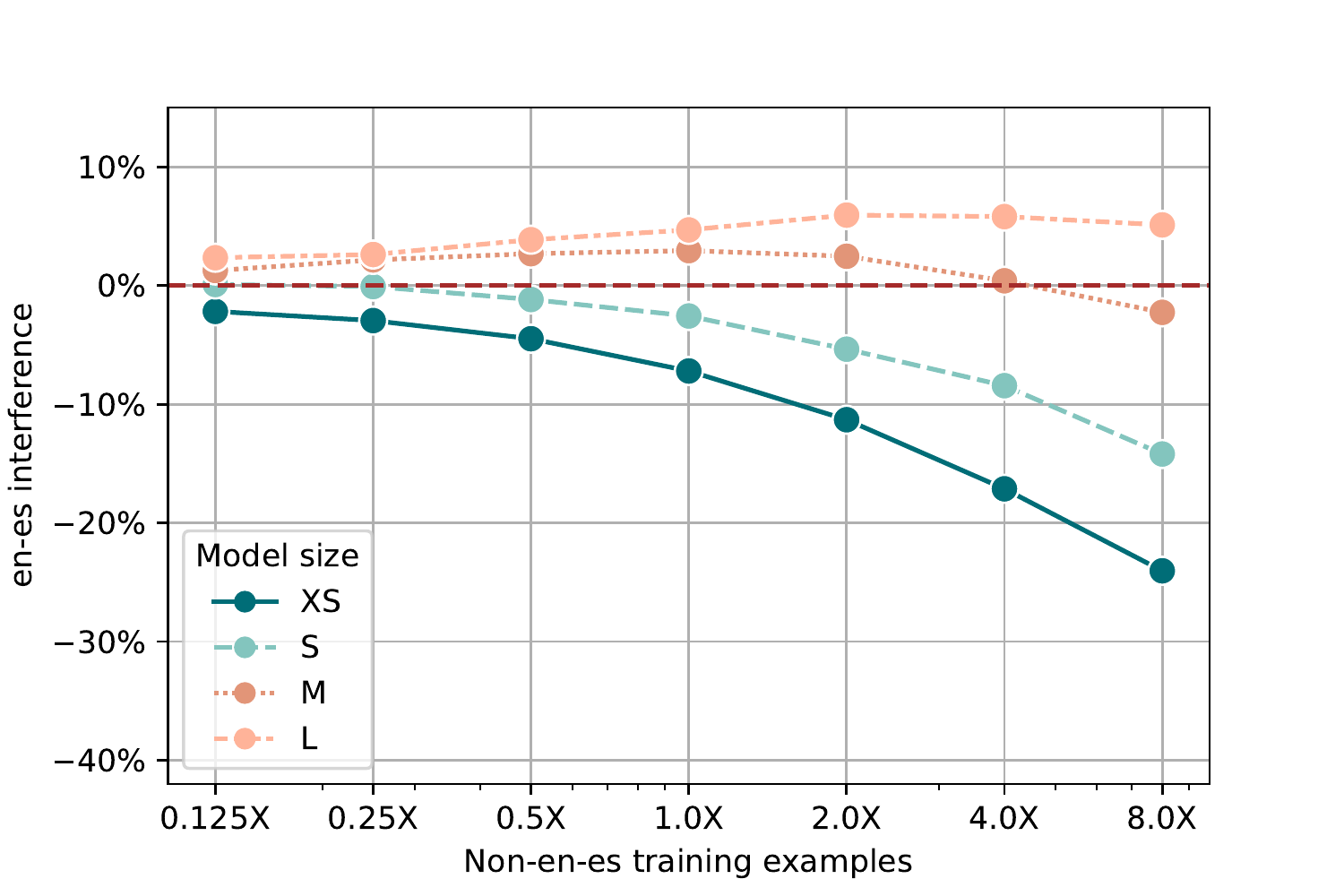}
  \caption{15.2M en-es examples}
  \label{fig:en_es_15.2M}
\end{subfigure}%
\begin{subfigure}{.5\textwidth}
  \centering
  \includegraphics[width=\columnwidth]{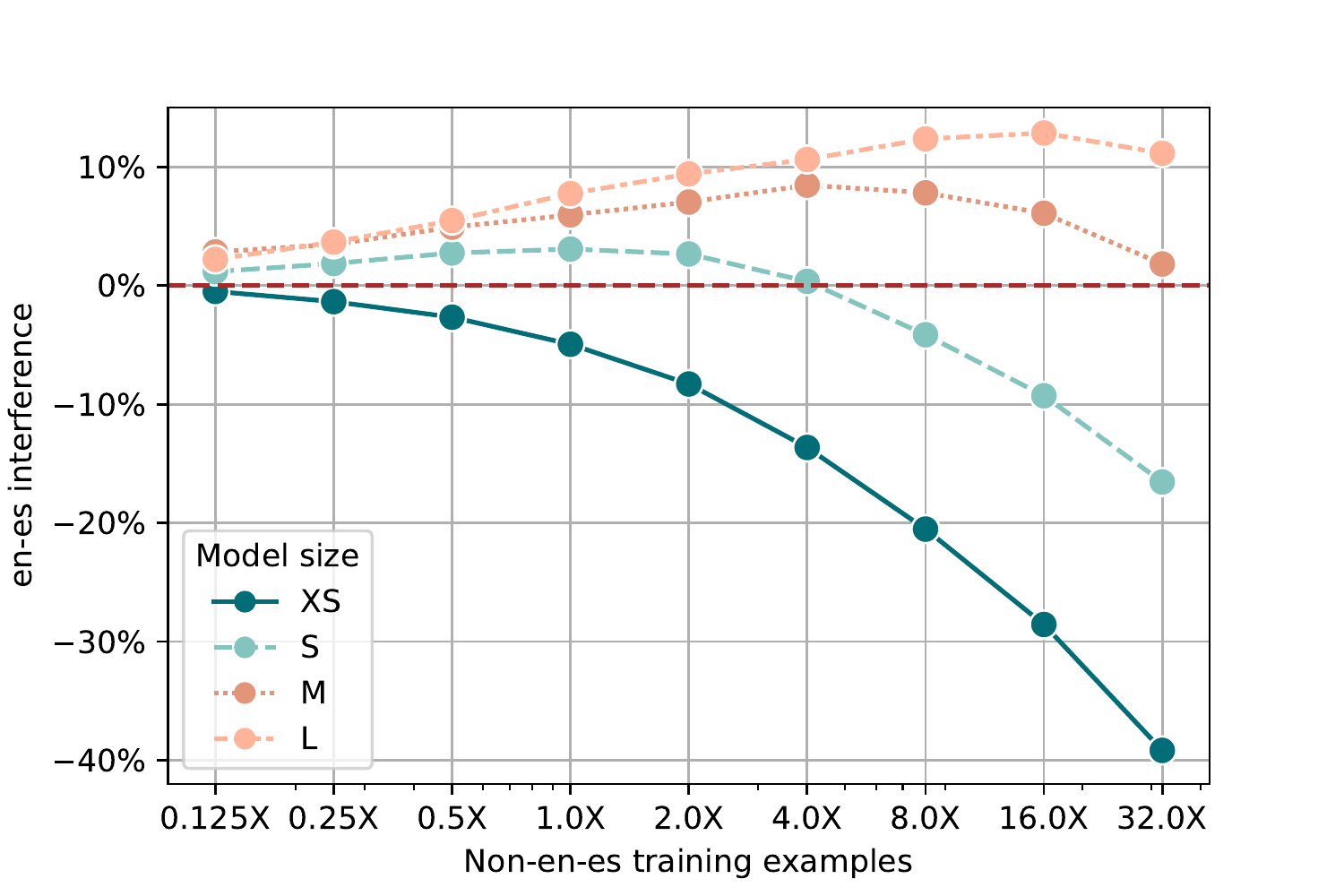}
  \caption{3.8M en-es examples}
  \label{fig:en_es_3.8M}
\end{subfigure}
\begin{subfigure}{.5\textwidth}
  \centering
  \includegraphics[width=\columnwidth]{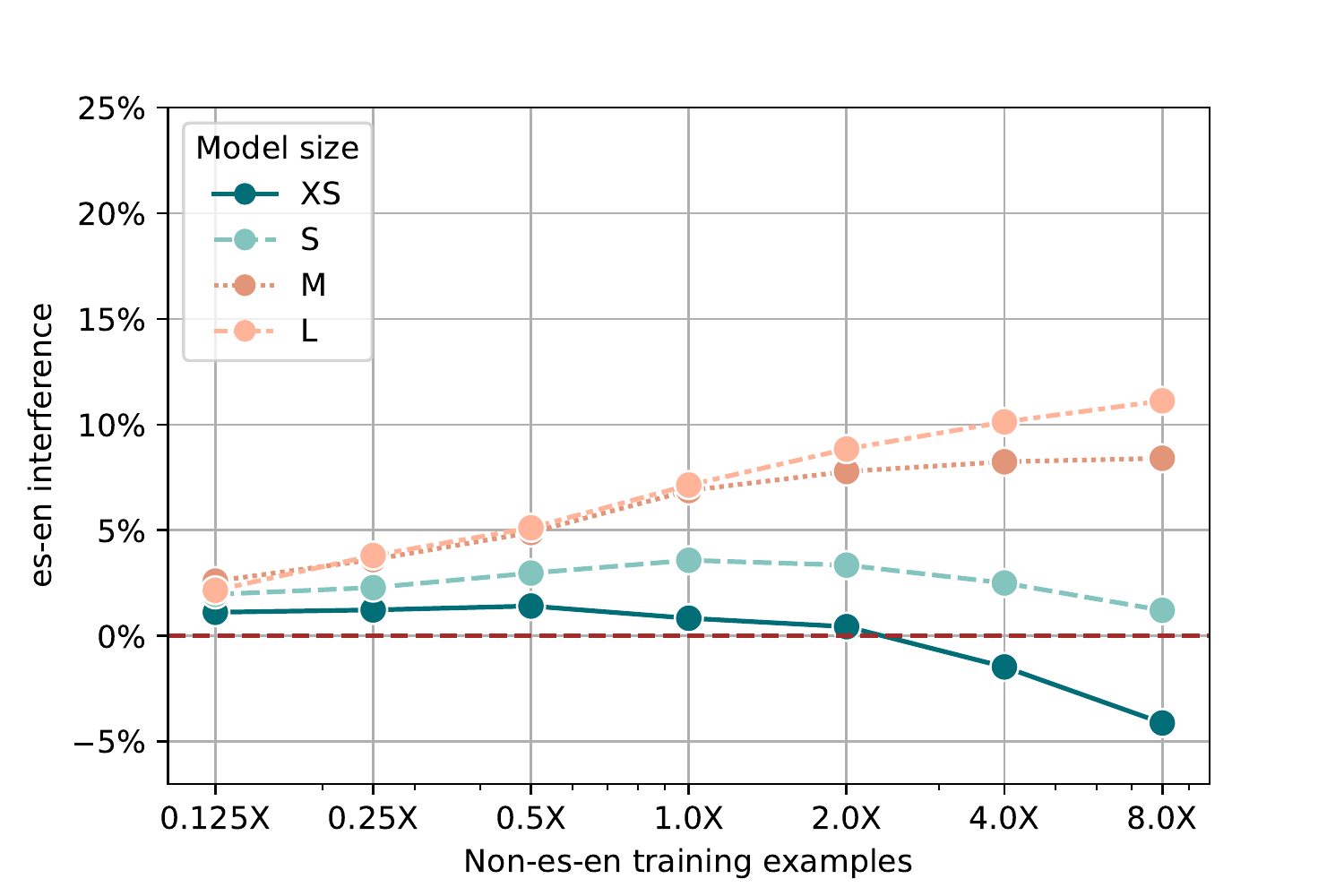}
  \caption{15.2M es-en examples}
  \label{fig:es_en_15.2M}
\end{subfigure}%
\begin{subfigure}{.5\textwidth}
  \centering
  \includegraphics[width=\columnwidth]{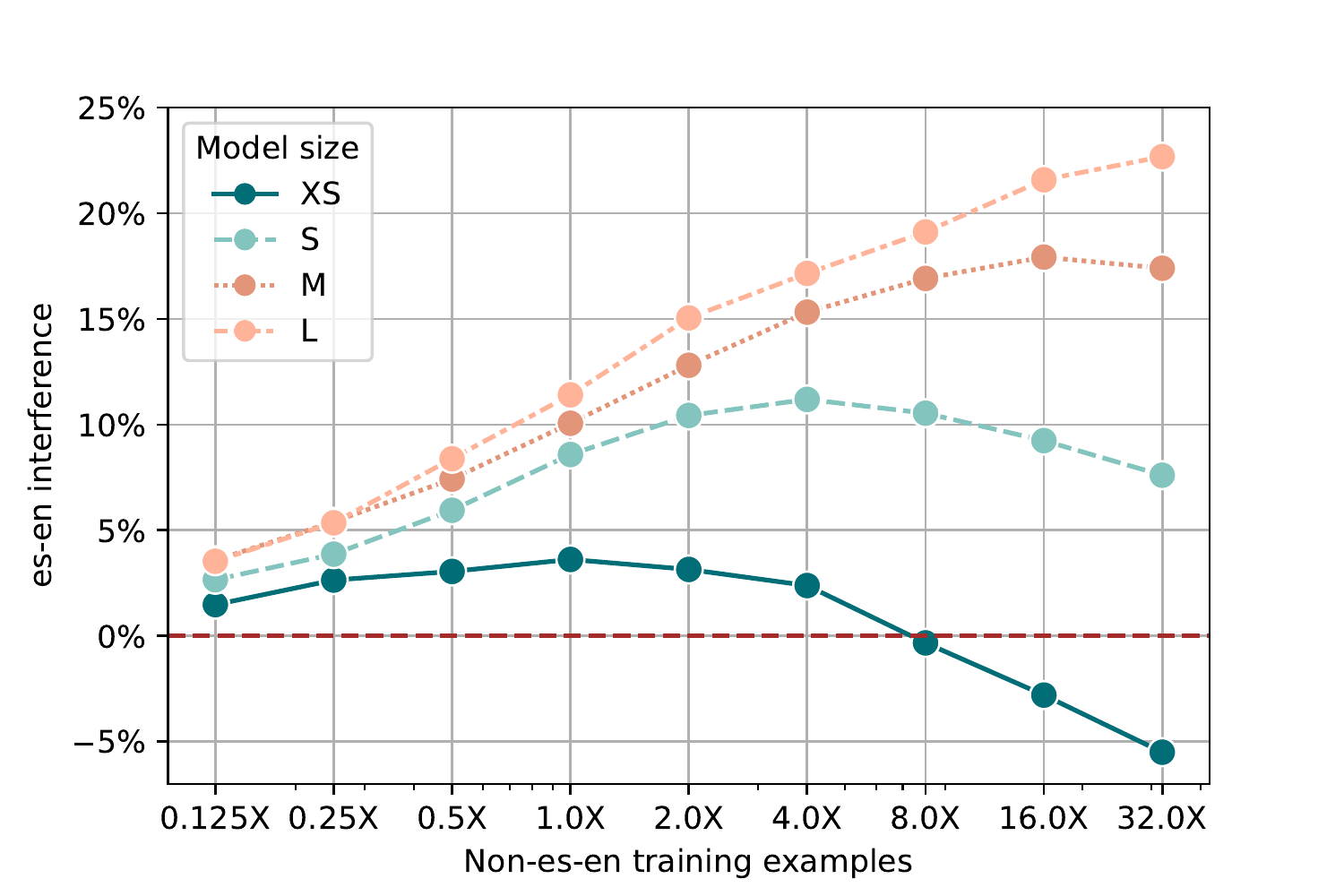}
  \caption{3.8M es-en examples}
  \label{fig:es_en_3.8M}
\end{subfigure}
\caption{Interference of en-es (top) and es-en (bottom) models trained using the full 15.2M en-es train set (left), and a sample of 3.8M en-es (right).  Positive values indicate synergy, i.e. en-es or es-en loss of a multilingual model is lower (better) comparing to its bilingual model baseline. Similarly, negative values indicate interference.
} 
\label{fig:interference_across_sizes}
\end{figure*}

\subsection{The Impact of Model and Data Size}
\label{sec:scale}
Seeing that language similarity and the number of interfering languages have only a limited effect on interference, we design a controlled setup to measure interference as a function of the remaining three factors: model size, focus language data size, and its proportion in the total amount of data seen during training.

\paragraph{Setup}
We train models using all the available 15.2M English-Spanish examples, with an increasing example budget for interfering language pairs, ranging from 1$/$8 (1.9M) to 8 times (122M) the  English-Spanish data, divided as evenly as possible between French, Czech, Russian, and Chinese.\footnote{Since Chinese has only 26M examples (less than 122M$/$4), we use all of its train set in the 122M (8.0X) case, and sample the remainder of the example budget from the three from French, Czech, and Russian.}
To observe trends across $D_{s \rightarrow t}$ sizes, we rerun these experiments with a quarter (3.8M) of the English-Spanish data, while keeping the ratios with the rest of the data similar.
Finally, we also conduct these experiments in the many-to-English setting.

\paragraph{Results}
Figures \ref{fig:en_es_15.2M} and \ref{fig:en_es_3.8M} show the interference and synergy for English-Spanish using a varying number of interfering examples.
For smaller models (XS and S), increasing the amount of interfering data (i.e. decreasing the proportion of focus data) exacerbates interference.
However, larger models appear to benefit from significant quantities of interfering examples; for instance, when training with $D_{s \rightarrow t}=$ 3.8M, a large model (L) can gain over 10\% relative loss improvement when there is 32 times more interfering data than focus data ($P(x \in s \rightarrow t) \approx 3\%$).
Interestingly, we also observe that interference is sensitive to the ratio between model parameters and focus data, as the M model trained on 15.2M focus examples produces a similar curve to that of the 4-times smaller S model trained on 3.8M examples, both intersecting the synergy/interference line at the same point.
Finally, Figures \ref{fig:es_en_15.2M} and \ref{fig:es_en_3.8M} show that when translating \textit{into} English, interference is much less of an issue, occurring only in the XS model when the total amount of training data significantly exceeds the model's capacity.
Scaling up the model not only improves the absolute performance (Appendix \ref{sec:appendix_bleu}), but also introduces substantial gains from synergy. 
Our results align with trends observed on cross lingual transfer when scaling pretrained multilingual models to 3.5 and 10 billion parameters \cite{goyal-etal-2021-larger}.

\subsection{Tuning Interference with Temperature}
In the previous sections we demonstrated that the dominant factors impacting interference are the model size, the amount of focus language pair data $D_{s \rightarrow t}$, and the proportion of focus pair examples observed during training $P(x \in s \rightarrow t)$.
In a practical situation where both model size and multilingual data are fixed, how can one control the level of interference?
Recalling Equation~\ref{eq:proportion}, we observe that the proportion of focus pair examples $P(x \in s \rightarrow t)$ is controlled via the temperature hyperparameter $T$.
Although previous literature has largely used a value of $T=5$ following \citet{Arivazhagan2019MassivelyMN}, our systematic experiments with different temperatures across three different data distributions and four model sizes suggest that this value can be sub-optimal and induce a substantial amount of interference, especially for model sizes that alleviate significant amounts of interference (M and L).
Conversely, tuning the temperature shows that lower values ($T=1,2$) are typically able to reduce high-resource interference without harming low-resource synergy in our standard multilingual translation setting.

\begin{figure}[t!]
\centering
\begin{subfigure}[b]{.48\textwidth}
\centering
   \includegraphics[width=1\linewidth]{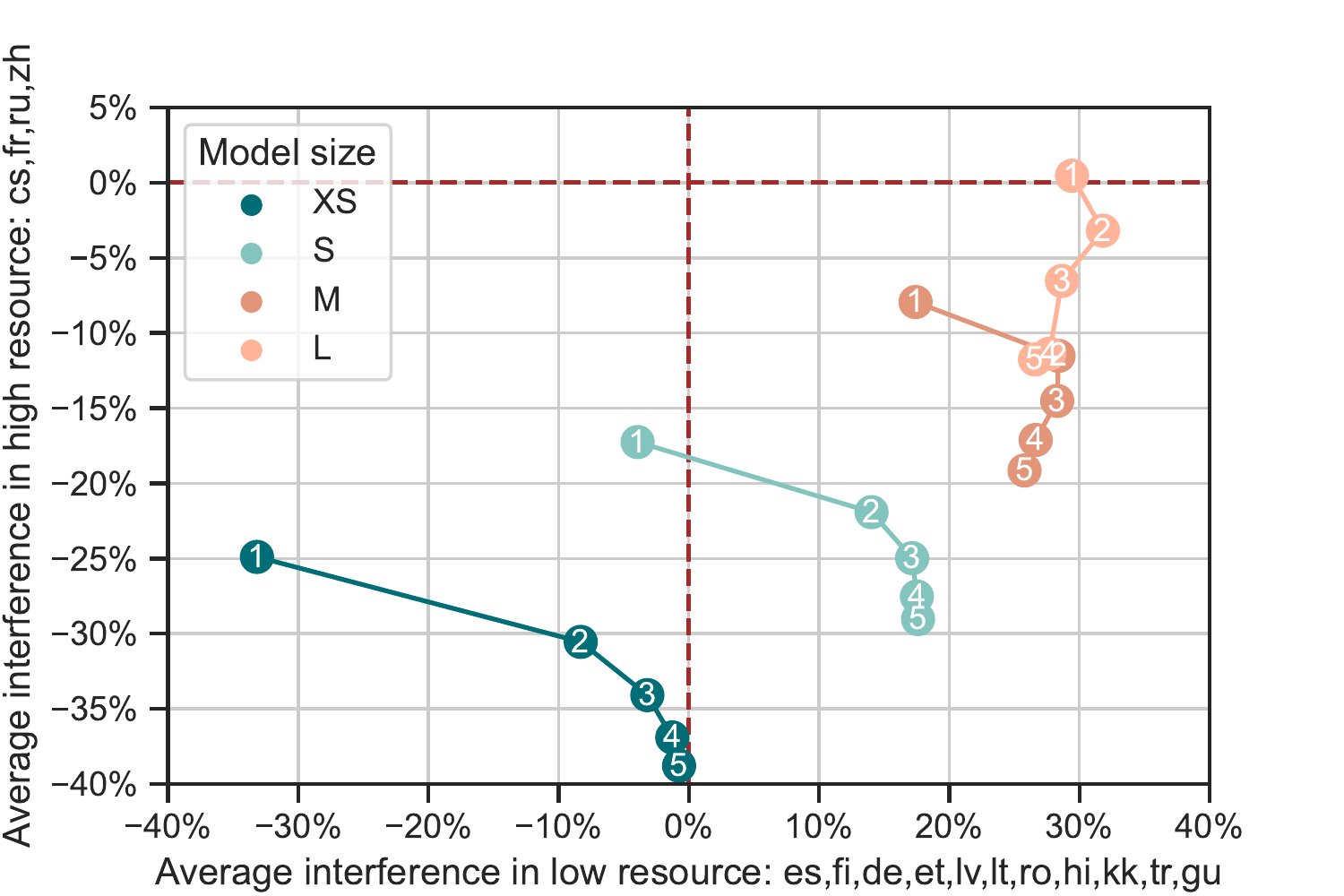}
  \caption{Trained on all languages}
   \label{fig:tmp_full_data} 
\end{subfigure}

\begin{subfigure}[b]{0.48\textwidth}
\centering
   \includegraphics[width=1\linewidth]{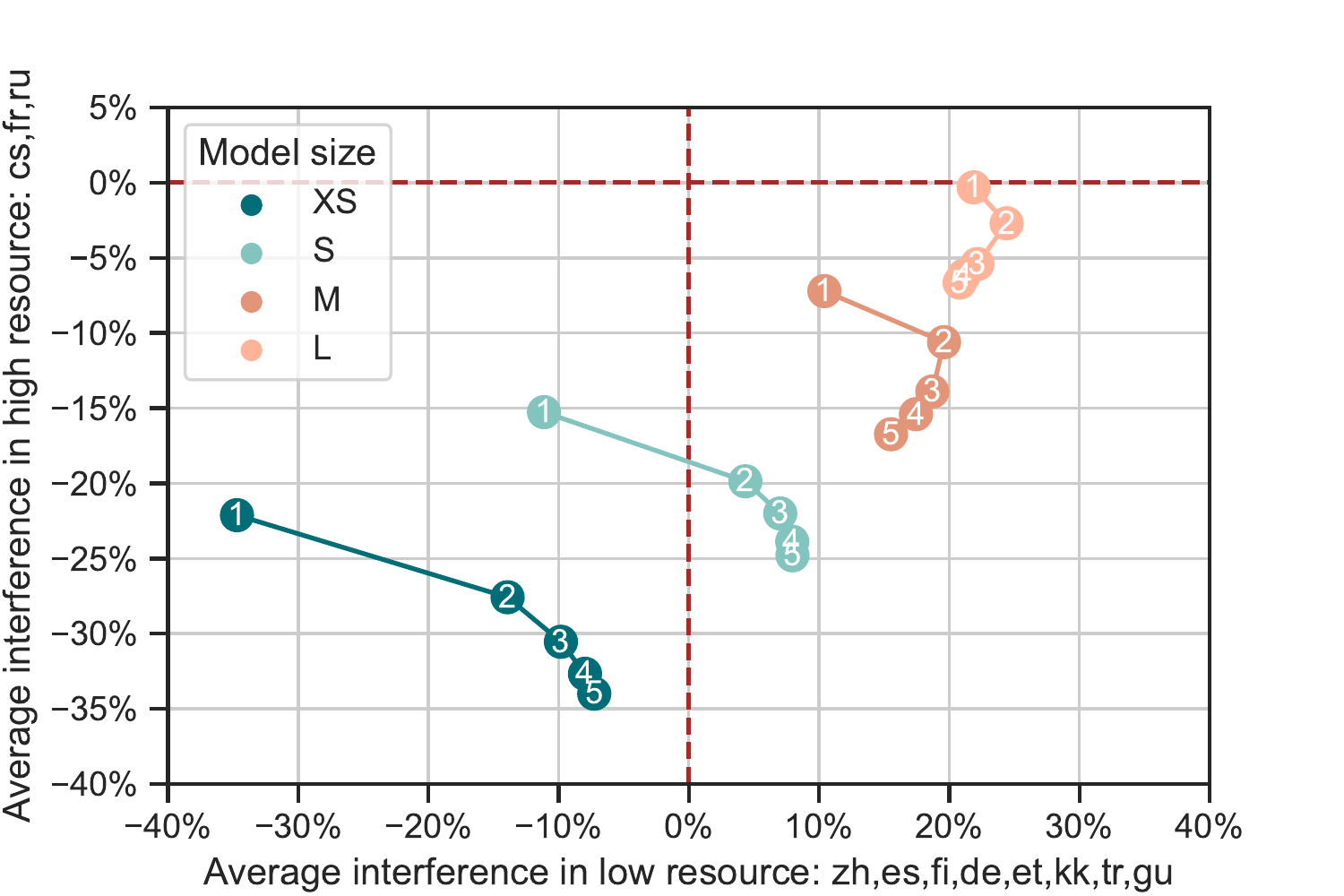}
  \caption{Trained w/o 4 low resource languages}
   \label{fig:tmp_few_low}
\end{subfigure}

\begin{subfigure}[b]{0.48\textwidth}
\centering
   \includegraphics[width=1\linewidth]{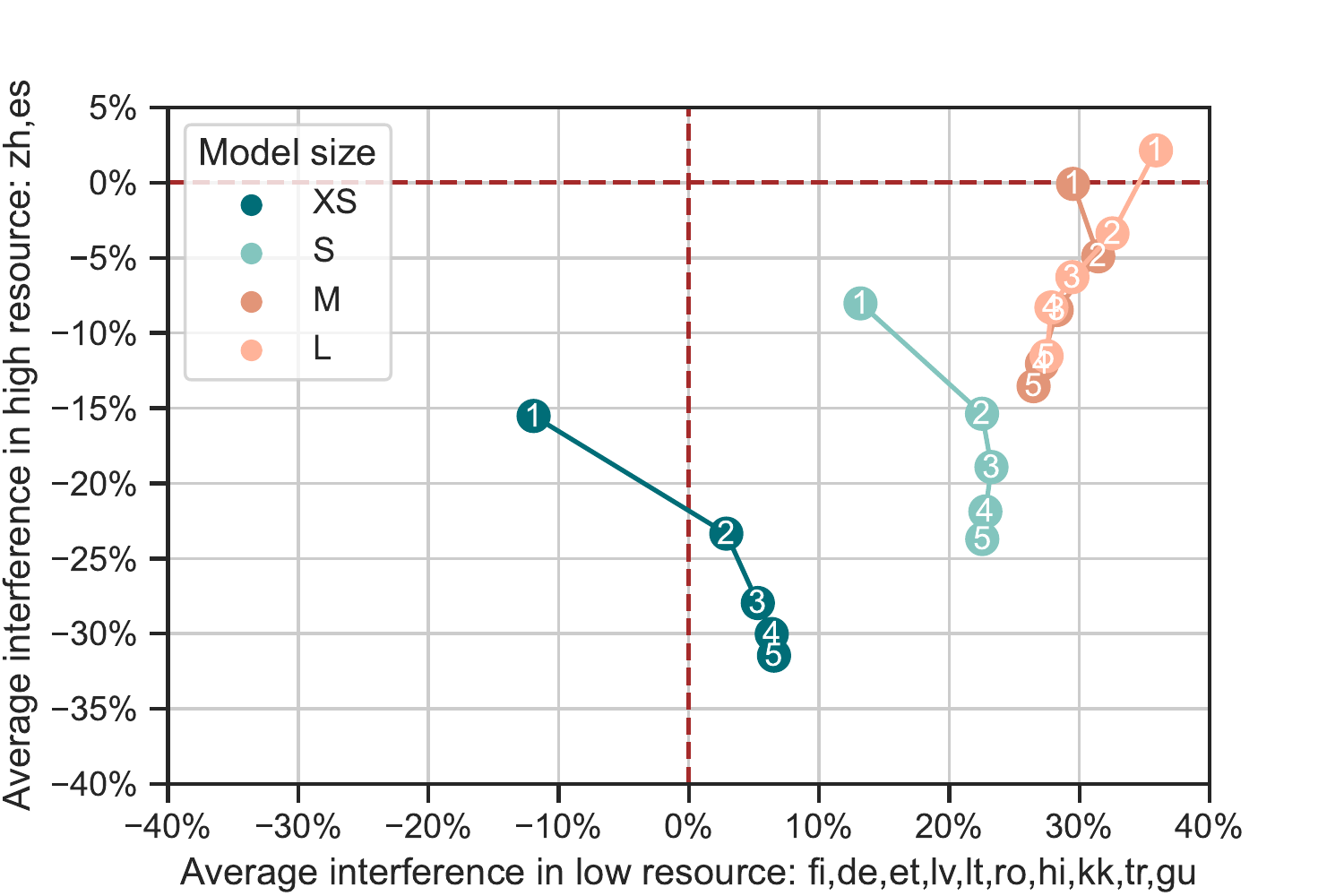}
  \caption{Trained w/o 3 high resource languages}
   \label{fig:tmp_few_high}
\end{subfigure}
\caption{Average interference/synergy of high (proportion declining when incrementing $T$) and low (proportion ascending when incrementing $T$) resource languages of different model sizes (colors) for different training distributions (a,b,c) using $T$ values ranging from 1 to 5 (numbers on markers). Positive values indicate synergy and negative values indicate interference.
}
\label{fig:tmp}
\end{figure}

\paragraph{Setup}
We train models of four sizes with temperature ranging from 1 to 5 on three training distributions: (1) all available training data, (2) discarding 3 high resource languages (Czech, French and Russian), (3) discarding 4 low resource languages (Latvian, Lithuanian, Romanian and Hindi). When illustrating the results, we assign languages to high and low resource according to whether their relative data proportion decreases or increases when going from $T=1$ to $T=2$.

\paragraph{Results}
Figure~\ref{fig:tmp} shows the trade-offs between the lower and higher resource languages, as defined above.
First, we can see a clear trade-off for the smaller models (XS and S) from $T=1$ to $T=4$ in most cases. Increasing $T$ helps promote synergy for low resource languages at the cost of increasing interference for the high resource languages.
However, the larger models (M and L) clearly degrade when using $T \geq 3$; in fact, values of $T=1$ and $T=2$ are often better for high- and low-resource language pairs than the commonly-used $T=5$.
These results align with recent work \citet{xin2022do} showing that tuned scalarization is key to achieving strong bilingual baselines that often outperform more complicated multitask optimization methods.\footnote{See Table \ref{tab:tmp_results} in Appendix \ref{sec:appendix_bleu} for the results of these experiments with absolute BLEU scores.}

\section{Related Work}

\paragraph{Scaling Laws in Machine Translation}
Previous work also looked at scaling trends of data and models sizes for machine translation. \citet{gordon-etal-2021-data} proposed scaling laws in the data and model parameters and demonstrated their ability to predict the validation loss of bilingual translation models from Russian, Chinese, and German to English. \citet{ghorbani2022scaling} found scaling laws for different configurations for the encoder and decoder, independently varying the number of layers in each of them. \citet{pmlr-v162-bansal22b} examined different architectures and described data size scaling laws for machine translation in a large scale for English to German and English to Chinese. While all of these works focused on the bilingual setting, we unveil trends for multilingual translation, which has increased complexity. Concurrently to our work, \citet{fernandes2023scaling} proposed scaling laws for multilingual machine translation, focusing on trilingual models trained on English-German with English-Chinese or French

\paragraph{Multitask Methods for Multilingual Machine Translation}
Multitask methods have been proposed extensively to enhance the performance of multilingual translation models. Some utilize validation based signals to determine which language pairs should be prioritized throughout training, either with adaptive scheduling \cite{Jean2019AdaptiveSF}, gradient similarities to the validation set \citet{wang-etal-2020-balancing}, or a multi-armed bandits model \cite{kreutzer-etal-2021-bandits-dont}.  \citet{zhu-etal-2021-counter-interference} added dedicated embedding and layer adapter modules to the Transformer, and \citet{lin-etal-2021-learning} suggested learning a binary mask for every model parameter and every language pair, both requiring further training after the base multilingual model converges.
\citet{li2021robust} used per language gradients geometry to rescale gradients of different language pair to improve performance on low resource languages. 
\citet{wang2021gradient} extended PCGrad \cite{NEURIPS2020_3fe78a8a} to create Gradient Vaccine, a method that attempts to deconflict different language pairs gradients by replacing them with more similar vectors in terms of cosine similarity. While the motivation for these methods is clear and intuitive, they are usually more complex and  computationally expensive than the baseline. Moreover, their efficacy is often demonstrated using relatively small\footnote{Transformer-base or big from \citet{vaswani2017attention}.} models, while modestly increasing the model size can both strengthen the bilingual baselines and reduce the interference problem significantly.

\paragraph{Critical Takes on Multitask Optimization Methods}
Multitask optimization methods were recently under scrutiny.
\citet{kurin2022in} experimented with many of those for image classification and reinforcement learning problems, and found that none of them consistently outperformed a well tuned baseline with proper use of known regularization techniques. Similarly, \citet{xin2022do} showed that despite their increased complexity, no popular multitask method was superior to a sweep over scalarization weights for a baseline trilingual translation model. This work complements this line of research by examining \textit{multilingual} translation models and how can modest scale and calibrated temperature reduce problems associated with multitasking.

\section{Conclusion}

This work examines the dominant factors that influence interference in multilingual machine translation. Namely, the model size, the amount of parallel data for the focus language pair, and the proportion of examples from the focus language pair with respect to the total data seen during training.
While specialized multitask techniques are sometimes demonstrated on small transformer models, we find that a standard baseline model of 176M parameters reduces the interference problem significantly, and further scaling up results in synergy among the different language pairs.
We further demonstrate the importance of tuning the temperature at which different language pairs are sampled during training; while existing literature largely relies on high temperatures, which indeed improve low-resource performance in parameter-poor settings, larger models benefit from a more natural distribution that reflects the raw training data.
These simple strategies for addressing interference call into question the necessity and perhaps even the validity of recently-proposed complex anti-interference methods and reaffirm the tried-and-true method of increasing model capacity to accommodate for higher data diversity.

\section{Limitations}
One limitation of this work is the focus on English-to-many and many-to-English settings, while previous studies also went beyond English-centric translation \cite{freitag-firat-2020-complete,10.5555/3546258.3546365}.  
Second, we experiment with a WMT based benchmark that has a total of 15 languages and ~200M training examples, when translation models were also trained on larger datasets \cite{aharoni-etal-2019-massively,Arivazhagan2019MassivelyMN,https://doi.org/10.48550/arxiv.2207.04672}.
We leave questions about the amount of scale that will be required to effectively mitigate interference in massively (many-to-many, billions of parallel sequences) multilingual settings for future work.
Additionally, the data collected from high resource languages may be of higher quality compared to that collected from low resource languages. Further research is needed to determine the impact of low quality training data on interference and synergy.
Finally, while we explore trends when scaling models width, deeper models \cite{ghorbani2022scaling} might help mitigating interference even further.

\section*{Acknowledgments}

 This research is supported by the Yandex Initiative in Machine Learning. We thank Maor Ivgi, Yilin Yang, Jean Maillard, and Ves Stoyanov for their valuable feedback.

\bibliography{anthology,custom}
\bibliographystyle{acl_natbib}

\appendix

\section{BLEU Scores}
\label{sec:appendix_bleu}

Throughout the paper we calculate interference in terms of test loss values. We additionally provide the test BLEU scores achieved by our models. We generate using beam search with 5 beams, without length penalty.
We use SacreBLEU \cite{post-2018-call} to calculate test sets BLEU \cite{papineni-etal-2002-bleu} scores.

\begin{figure*}[t]
\centering
\begin{subfigure}[b]{1.0\textwidth}
\centering
   \includegraphics[width=1\linewidth]{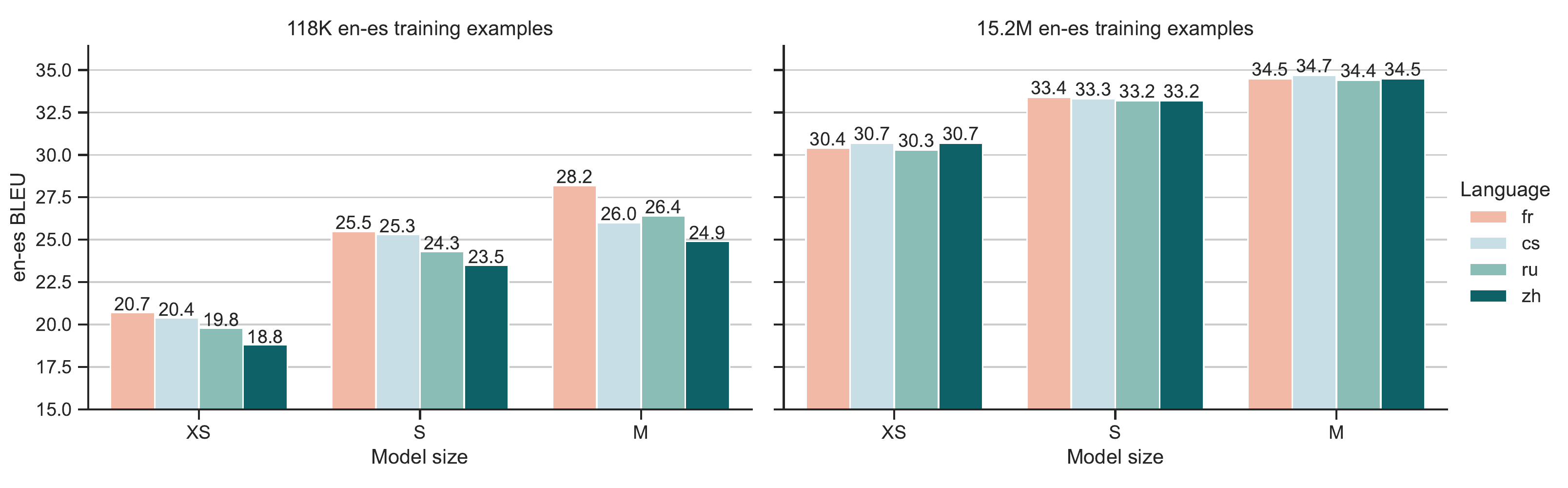}
   \caption{Models trained with 118K (left) and 15.2M (right) en-es training examples and 15.2M training examples for non-es languages. }
   \label{fig:abs_bleu_lang_sim_es} 
\end{subfigure}

\begin{subfigure}[b]{1.0\textwidth}
\centering
   \includegraphics[width=1\linewidth]{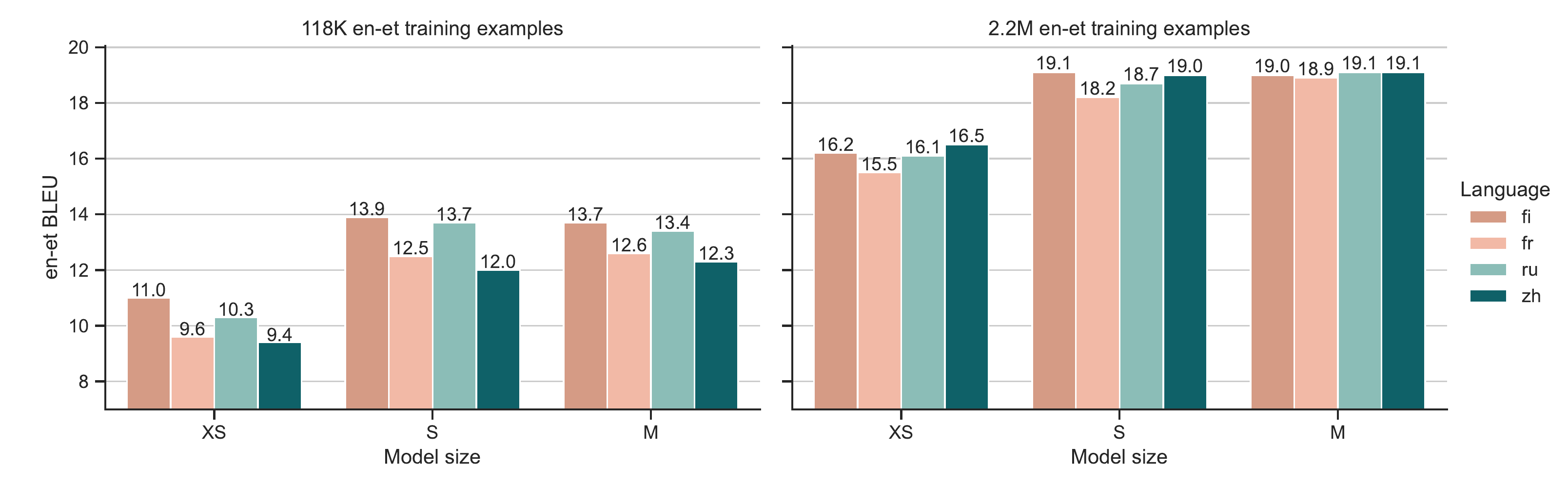}
   \caption{Models trained with 118K (left) and 2.2M (right) en-et training examples and 6.6M training examples for non-et languages.}
   \label{fig:abs_bleu_lang_sim_et}
\end{subfigure}

\caption{en-es (a) and en-et (b) test BLEU scores of models trained with es or et as low resource languages and using their full train sets together with one other en-xx pair.
}
\label{fig:abs_bleu_lang_sim}
\end{figure*}

\paragraph{Language similarities}
Figure~\ref{fig:abs_bleu_lang_sim} shows BLEU scores of models from experiments in Section~\ref{sec:lang_sim}. They reflect similar trends, as the variance between different interfering languages when the focus language has only 118K examples diminish when a decent amount of training data is available.

\begin{figure*}[t]
\centering
\begin{subfigure}[b]{1.0\textwidth}
\centering
   \includegraphics[width=1\linewidth]{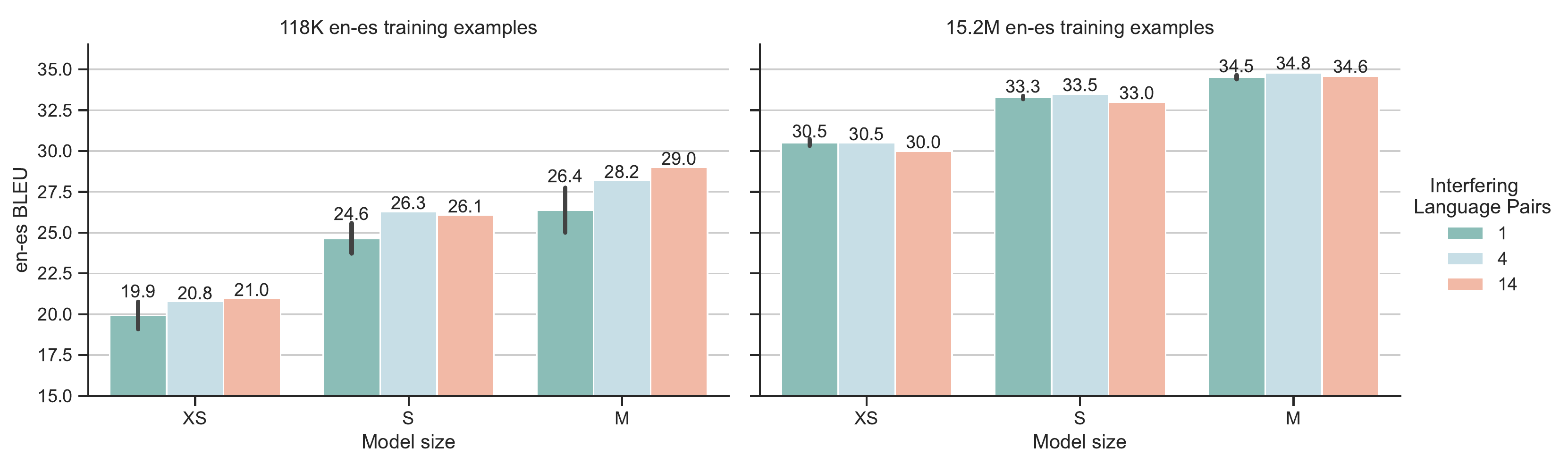}
   \caption{Models trained with 118K (left) and 15.2M (right) en-es training examples and 15.2M training examples for non-es languages. }
   \label{fig:abs_bleu_num_tasks_es} 
\end{subfigure}

\begin{subfigure}[b]{1.0\textwidth}
\centering
   \includegraphics[width=1\linewidth]{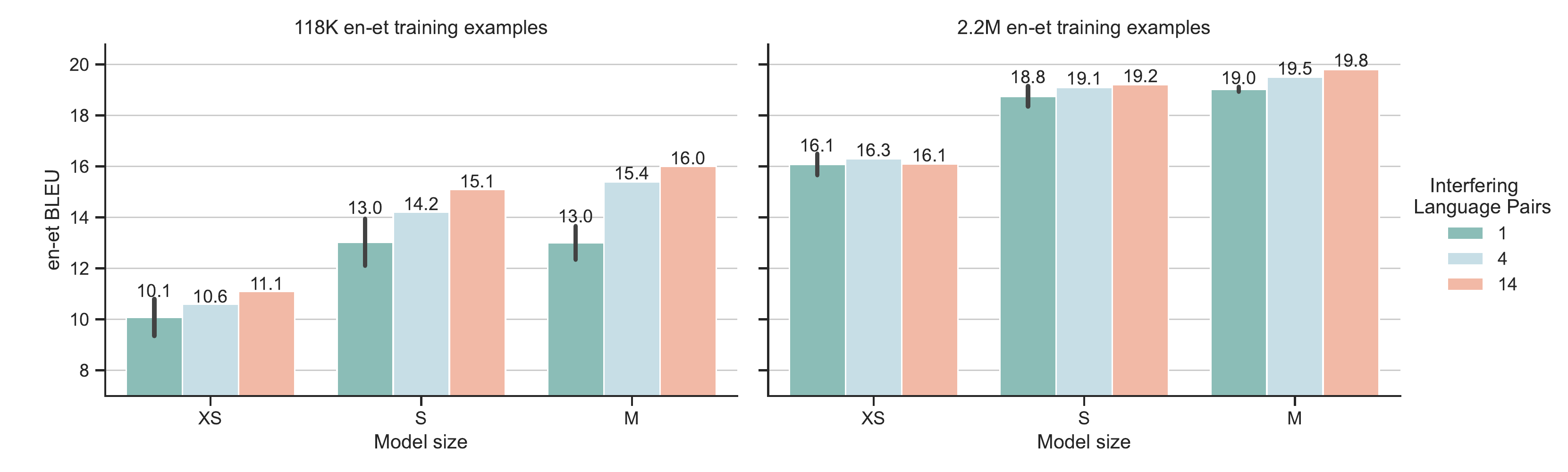}
   \caption{Models trained with 118K (left) and 2.2M (right) en-et training examples and 6.6M training examples for non-et languages.}
   \label{fig:abs_bleu_num_tasks_et}
\end{subfigure}

\caption{en-es (a) and en-et (b) test BLEU scores of models trained with es or et as low resource languages and using their full train sets together with increasing number of languages, sharing a fixed budget of training examples.}
\label{fig:abs_bleu_num_tasks}
\end{figure*}

\paragraph{Number of languages}
Figure~\ref{fig:abs_bleu_num_tasks} shows BLEU scores of models from experiments in Section~\ref{sec:num_tasks}. They also demonstrate that low resource pairs benefit when there are more interfering languages, but this effect disapper with a decent amount of training data.

\begin{table*}[hb!] 
\small
\centering
\begin{tabular}{@{}lcccccccccccccccc@{}}
\toprule
 \textbf{Size} & \textbf{Tmp} &\textbf{cs} & \textbf{fr} & \textbf{ru} & \textbf{zh} & \textbf{es} & \textbf{fi} & \textbf{de} & \textbf{et} & \textbf{lv} & \textbf{lt} & \textbf{ro} & \textbf{hi} & \textbf{kk} & \textbf{tr} & \textbf{gu} \\
\midrule
\multirow{6}{*}{XS}  & bi & \textbf{19.6} & \textbf{35.1} & \textbf{24.2} & \textbf{27.3} & \textbf{31.7} & \textbf{17.7} & \textbf{24.1} & \textbf{17.5} & 12.1 & 9.2 & \textbf{22.4} & 6.5 & 0.5 & 7.7 & 1.6 \\
  & 1 & 16.7 & 31.9 & 20.5 & 20.8 & 27.4 & 12.8 & 15.7 & 10.4 & 8.0 & 6.1 & 15.6 & 4.8 & 1.0 & 4.9 & 2.5 \\
  & 2 & 15.5 & 30.8 & 19.0 & 20.3 & 27.4 & 14.1 & 17.6 & 13.1 & 11.3 & 9.4 & 20.4 & 9.1 & 2.3 & 9.1 & 6.4 \\
  & 3 & 15.2 & 30.2 & 18.6 & 19.6 & 27.3 & 14.6 & 18.0 & 13.5 & 11.9 & 9.8 & 21.5 & 11.0 & 3.2 & 10.2 & 7.6 \\
  & 4 & 14.8 & 30.1 & 18.2 & 19.4 & 27.1 & 14.7 & 18.1 & 13.7 & 12.4 & 10.0 & 21.5 & 11.7 & 3.2 & 10.7 & 8.7 \\
  & 5 & 14.5 & 29.9 & 17.6 & 19.0 & 27.1 & 14.6 & 18.1 & 13.6 & \textbf{12.6} & \textbf{10.3} & 21.7 & \textbf{11.9} & \textbf{3.5} & \textbf{10.8} & \textbf{9.2} \\
\midrule
\multirow{6}{*}{S}  & bi & \textbf{22.1} & \textbf{38.4} & \textbf{27.2} & \textbf{29.9} & \textbf{33.8} & \textbf{19.8} & \textbf{26.1} & 17.4 & 12.0 & 8.5 & 22.1 & 4.8 & 0.5 & 7.2 & 1.8 \\
  & 1 & 20.3 & 36.2 & 24.7 & 26.4 & 31.0 & 16.8 & 20.8 & 14.5 & 12.1 & 9.8 & 21.0 & 7.9 & 1.7 & 7.6 & 4.9 \\
  & 2 & 19.9 & 35.7 & 24.1 & 25.7 & 31.4 & 18.5 & 22.4 & 17.3 & 14.9 & 12.2 & 24.1 & 14.1 & 4.6 & 12.1 & 11.6 \\
  & 3 & 19.2 & 35.6 & 23.5 & 25.6 & 31.2 & 18.4 & 22.5 & 17.6 & 15.3 & 12.5 & 24.5 & 15.2 & 5.6 & 12.9 & 13.1 \\
  & 4 & 19.1 & 35.2 & 23.7 & 25.0 & 30.9 & 17.5 & 22.6 & \textbf{17.7} & \textbf{15.4} & \textbf{12.8} & \textbf{25.0} & 15.3 & 5.7 & 13.3 & 13.1 \\
  & 5 & 18.5 & 34.8 & 23.4 & 25.1 & 30.9 & 18.1 & 22.3 & 17.5 & 15.3 & 12.5 & 24.9 & \textbf{15.4} & \textbf{5.9} & \textbf{13.8} & \textbf{13.5} \\
\midrule
\multirow{6}{*}{M}  & bi & \textbf{23.1} & \textbf{40.1} & \textbf{28.8} & \textbf{30.7} & \textbf{34.2} & 19.6 & 25.9 & 17.1 & 11.5 & 7.8 & 21.6 & 4.0 & 0.4 & 5.9 & 1.0 \\
  & 1 & 22.4 & 39.6 & 27.3 & 29.8 & 33.6 & 19.1 & 24.1 & 18.0 & 14.6 & 12.0 & 23.9 & 12.4 & 3.6 & 10.7 & 8.2 \\
  & 2 & 22.1 & 39.3 & 26.5 & 29.7 & 33.5 & 19.5 & 25.7 & 19.3 & 17.1 & 13.8 & \textbf{26.5} & \textbf{15.9} & \textbf{6.3} & 14.1 & \textbf{14.2} \\
  & 3 & 21.8 & 38.0 & 26.1 & 29.6 & 33.4 & 20.1 & \textbf{26.1} & \textbf{20.2} & \textbf{17.4} & 13.8 & \textbf{26.5} & 15.2 & 5.8 & \textbf{14.2} & 14.1 \\
  & 4 & 21.3 & 38.0 & 25.9 & 29.0 & 33.4 & \textbf{20.3} & 25.8 & 20.1 & 16.9 & \textbf{14.1} & \textbf{26.5} & 14.6 & 5.5 & 13.7 & 12.2 \\
  & 5 & 21.1 & 37.7 & 26.2 & 28.6 & 32.8 & 19.9 & 25.6 & 19.4 & 16.8 & 13.9 & 26.3 & 14.6 & 5.2 & 13.8 & 12.3 \\
\midrule
\multirow{6}{*}{L}  & bi & 22.9 & 40.0 & 28.5 & 30.7 & 34.4 & 18.6 & 25.8 & 16.9 & 10.8 & 8.5 & 21.4 & 3.8 & 0.4 & 5.4 & 1.3 \\
  & 1 & \textbf{23.4} & \textbf{40.7} & \textbf{29.4} & \textbf{31.4} & 34.8 & 20.7 & 26.5 & 19.2 & 16.3 & 13.4 & 26.1 & \textbf{14.4} & 4.6 & 12.5 & 10.3 \\
  & 2 & 23.0 & 40.4 & 29.1 & 31.1 & 34.7 & 20.6 & \textbf{28.0} & 20.2 & \textbf{17.9} & \textbf{14.2} & \textbf{26.7} & 14.2 & \textbf{4.7} & \textbf{14.2} & 12.4 \\
  & 3 & 22.9 & 39.8 & 28.4 & 31.1 & \textbf{34.9} & \textbf{21.3} & 27.7 & \textbf{20.5} & 17.4 & \textbf{14.2} & 26.2 & 13.5 & 4.6 & 14.0 & 12.2 \\
  & 4 & 22.1 & 39.2 & 26.5 & 29.8 & 34.0 & 20.5 & 26.7 & 20.3 & 17.3 & \textbf{14.2} & 26.4 & 13.8 & \textbf{4.7} & 14.0 & 12.1 \\
  & 5 & 21.9 & 38.9 & 27.5 & 30.1 & 34.1 & 21.1 & 26.7 & 20.4 & 17.2 & 13.7 & 25.8 & 13.6 & 3.8 & 13.9 & \textbf{13.0} \\
\bottomrule
\end{tabular}
\caption{
Test BLEU scores across four model sizes of bilingual baselines (bi) and multilingual models trained with temperature values $T \in [1,5]$.}
\label{tab:tmp_results}
\end{table*}

\end{document}